\documentclass[lettersize,journal]{IEEEtran}

\usepackage{amsmath,amsfonts}
\usepackage{algorithm}
\usepackage{array}
\usepackage[caption=false,font=normalsize,labelfont=sf,textfont=sf]{subfig}
\usepackage{textcomp}
\usepackage{stfloats}
\usepackage{url}
\usepackage{verbatim}
\usepackage{graphicx}
\usepackage{bm}
\usepackage{cite}

\usepackage{algpseudocode}
\usepackage{xcolor}
\usepackage{hyperref}

\usepackage{multirow}
\usepackage{booktabs, caption}
\usepackage{tabularx}
\newcolumntype{C}{>{\centering\arraybackslash}X}
\newcommand\rebuttal [1]{\textcolor{black}{#1}}
\newcommand\newrebuttal [1]{\textcolor{black}{#1}}

\begin{document}

%\title{Overcoming Data Scarcity in Industrial NILM with Digital Twin-generated Dataset and Data Augmentation}
\title{Industrial Energy Disaggregation with Digital Twin-generated Dataset and Efficient Data Augmentation}

\author{%
\IEEEauthorblockN{%
  Christian Internò\IEEEauthorrefmark{1}\IEEEauthorrefmark{2},
  Andrea Castellani\IEEEauthorrefmark{2},
  Sebastian Schmitt\IEEEauthorrefmark{2},
  Fabio Stella\IEEEauthorrefmark{3}, and
  Barbara Hammer\IEEEauthorrefmark{1}%
}
% --- UPDATED AND REORDERED THANKS NOTES FOR CLARITY ---

% 1. Affiliations
\thanks{\IEEEauthorrefmark{1} C. Internò and B. Hammer are with the Machine Learning Group, Center for Cognitive Interaction Technology (CITEC), University of Bielefeld, Bielefeld, Germany.}
\thanks{\IEEEauthorrefmark{2} C. Internò, A. Castellani and S. Schmitt are with the Honda Research Institute EU, Offenbach am Main, Germany.}
\thanks{\IEEEauthorrefmark{3} F. Stella is with the Models and Algorithms for Data and Text Mining Laboratory (MADLab), Department of Informatics, Systems and Communication (DISCo), University of Milano - Bicocca, Milan, Italy.}

% 2. Other Notes (Now cleaner without symbols)
\thanks{Christian Internò and Andrea Castellani contributed equally to this work.}
\thanks{This work has been submitted to the IEEE for possible publication. Copyright may be transferred without notice, after which this version may no longer be accessible.}
}

% The paper headers
\markboth{PREPRINT VERSION - UNDER REVIEW}
{Internò \MakeLowercase{\textit{et al.}}: Industrial Energy Disaggregation with Digital Twin-generated Dataset}

\maketitle

\begin{abstract}
Industrial Non-Intrusive Load Monitoring (NILM) is limited by the scarcity of high-quality datasets and the complex variability of industrial energy consumption patterns. To address data scarcity and privacy issues, we introduce the Synthetic Industrial Dataset for Energy Disaggregation (SIDED), an open-source dataset generated using Digital Twin simulations. SIDED includes three types of industrial facilities across three different geographic locations, capturing diverse appliance behaviors, weather conditions, and load profiles. We also propose the Appliance-Modulated Data Augmentation (AMDA) method, a computationally efficient technique that enhances NILM model generalization by intelligently scaling appliance power contributions based on their relative impact. 
We show in experiments that NILM models trained with AMDA-augmented data significantly improve the disaggregation of energy consumption of complex industrial appliances like combined heat and power systems. Specifically, in our out-of-sample scenarios, models trained with AMDA achieved a Normalized Disaggregation Error of 0.167, outperforming models trained without data augmentation (0.451) and those trained with state-of-the-art data augmentation methods (0.290). Data distribution analyses confirm that AMDA effectively aligns training and test data distributions, enhancing model \rebuttal{generalization}. \rebuttal{The dataset SIDED and related code
is publicly available at \url{https://github.com/ChristianInterno/SIDED}.}
\end{abstract}


\section{Introduction}
\label{sec:introduction}
\IEEEPARstart{E}{nergy} management has become increasingly important due to the undeniable reality of climate change and the rising global energy demand \cite{IEA2021}. The industrial sector plays a significant role in international energy optimization \cite{492365,engel2024hierarchical}, necessitating heightened awareness of energy consumption to enhance efficiency and sustainability.

Non-Intrusive Load Monitoring (NILM) \cite{9820770} enables monitoring of individual appliance energy consumption without installing dedicated sensors, by analyzing measurements from the main power meter. This approach avoids the impracticality and expense of widespread sensor deployment \cite{8691975} and enhances the accuracy of energy management systems \cite{DBLP:journals/corr/abs-2105-00349}.

\rebuttal{While machine learning-based NILM methods have achieved results on residential datasets, where appliances generally exhibit discrete on/off behavior and low variability \cite{verma2021comprehensive,8818314}, these methods struggle in industrial environments. 
Industrial facilities include a greater variance and magnitude of power signals, with continuously varying, non-linear power profiles from Variable Frequency Drives (VFDs) and Constantly-on Variable Appliances (CVAs), along with integrated energy-producing systems.
In addition, the scarcity of publicly available industrial datasets, exacerbated by privacy and security issues \cite{IQBAL2021106921,10.1145/3447555.3464863,yan2023review,angelis2022nilm}, further complicates model development and generalization for industrial applications.}

While recent work has explored the use of Digital Twin technology to simulate industrial energy data \cite{100ffs32209,10226132,9179030,YU2022112407,9817461}, no comprehensive and fully accessible industrial NILM dataset has been produced that captures the full complexity of industrial loads. 
Moreover, conventional data augmentation (DA) techniques—designed primarily for residential data, are not suited to address the continuous overlapping nature of industrial signals and do not account for the wide variance in appliance power levels. 

\rebuttal{Our contributions are twofold:}

\rebuttal{\textbf{1)} SIDED, a novel Digital Twin-generated Synthetic Industrial Dataset for Energy Disaggregation. SIDED encompasses three types of industrial facility configurations across three geographic locations (Los Angeles, Tokyo, and Offenbach), capturing diverse appliance behaviors, weather conditions, and load profiles to provide a rich benchmark for industrial NILM research. 
SIDED leverages Digital Twin technology carefully calibrated on real operational data \cite{pub4012}, which is uniquely tailored to capture the intricate dynamics, heterogeneity, and non-stationarity of industrial energy consumption.}

\rebuttal{\textbf{2)} Appliance-Modulated Data Augmentation (\texttt{AMDA}), specifically designed to handle the continuous and overlapping nature of industrial energy signals. \texttt{AMDA} is designed to enrich the training dataset with context-specific data points by analyzing the incidence and power levels of each appliance within an industrial building. Other DA techniques \cite{iglesias2023data}, such as rotation, translation, cropping, flipping, or mixing do not adequately cover the variance in power magnitudes, which is critical for enhancing the \rebuttal{generalization} of industrial NILM systems due to the strong presence of CVA and VFD appliances. Our method significantly enhances the training data, facilitating the training of NILM models in a targeted, non-random manner. \texttt{AMDA} is particularly efficient in resource-constrained NILM environments, thanks to its modulated augmentation strategy.}

\rebuttal{We validate the effectiveness of \texttt{AMDA} through two NILM evaluations using the SIDED dataset and state-of-the-art models.
The first evaluation works as proximation when the system’s adaptability to appliance-level changes, such as reconfiguration or removal.
The second focuses on the model’s generalization across different facility setups, characterized by varying weather conditions and load profiles.
Additionally, we analyze how \texttt{AMDA} improves the alignment between training and test data distributions.
Together, these evaluations demonstrate \texttt{AMDA}’s ability bridge distribution gaps in industrial scenarios.}

\rebuttal{The rest of the paper is structured as follows:
in Section~\ref{sec:related}, we review related works in NILM, highlighting the gaps in the current literature.
%key differences between residential and industrial applications. 
Section~\ref{sec:comparison} outlines the general NILM problem and provides a comparative analysis of industrial versus residential NILM techniques, illustrating how different load characteristics necessitate distinct models. 
In Section~\ref{sec:dataset}, we introduce the SIDED dataset and detail the proposed Appliance-Modulated Data Augmentation (\texttt{AMDA}) approach. 
In Section~\ref{sec:exp}, we outline the experimental setup used in our work, then Section~\ref{sec:experiment} presents quantitative results on both appliance variation and facility variation scenarios. 
Finally, Section~\ref{sec:conclusion} concludes the paper and discusses future directions.}

\section{Related Work}
\label{sec:related}
While traditional Non-Intrusive Load Monitoring (NILM) techniques like Factorial Hidden Markov Models are effective for residential appliances with discrete on/off states \cite{7409513}, they struggle in industrial settings where VFD and CVA devices exhibit smooth and overlapping power signals \cite{angelis2022nilm, 10.1145/3447555.3464863}. Such continuous load variations require more flexible models capable of handling non-linear dynamics.

Classical ML methods, such as SVMs and Decision Trees \cite{7409513}, have been applied to NILM, but  Deep Neural Networks (DNNs) are preferred for their accuracy \cite{verma2021comprehensive}. However, DNNs require large labeled datasets, which are challenging to obtain in industrial settings due to limited data, privacy issues, and diverse equipment \cite{yan2023review}.

Data scarcity in industrial NILM settings is compounded by the strict confidentiality requirements of industrial processes, the unique operational schedules of each facility, and generally higher power levels than those seen in residential contexts \cite{10.1145/3447555.3464863, 9820770, angelis2022nilm}. To address the lack of annotated data, researchers have proposed DA strategies adapted from time-series analysis. \newrebuttal{These include techniques like Random Scaling \cite{Delfosse2020DeepLA}, adding random noise (Jitter) \cite{Wan2024Nonintrusive}, temporal shifting (Shift) \cite{Wan2024Nonintrusive}, and modifications in the frequency domain (FreqDropout) \cite{Wan2024Nonintrusive}, as well as more specialized methods for NILM like Denton-Cholette Interpolation \cite{LeQuy2022Data} , OFFSETAUG \cite{Francou2023Expanding}, isolates appliance ’on’ activations and adds a random constant offset to the high-power
portions of the signal, and InterUser \cite{Yang2025Intra}, creates new training samples by mixing appliance data from different facilities or ”users”.} However, these approaches must be applied cautiously to avoid introducing artifacts that misrepresent actual consumption behavior \cite{angelis2022nilm}. Moreover, many DA techniques have been tailored for residential appliances and do not directly generalize to continuously varying industrial loads \cite{9817461,100ffs32209}.

Another avenue for expanding training resources is the use of Generative Adversarial Networks (GANs) \cite{9426917}, which can synthesize realistic consumption patterns. GAN-based solutions, however, often require large amounts of pre-existing data and computational power \cite{8720065}, presenting further barriers in resource-constrained industrial settings. 

Digital Twin technology has recently emerged as a promising tool for generating synthetic data that closely mirrors real-world industrial conditions, while maintaining privacy \cite{8720065,10021651,YU2022112407}. Although Digital Twins have been used for tasks like anomaly detection \cite{9179030}, smart grid optimization \cite{CespedesCubides2024}, and industrial IoT energy management \cite{10226132}, their application in producing comprehensive synthetic datasets for NILM remains limited. Much of the existing work centers on operational optimization rather than generating load signatures of multiple industrial devices. Additionally, there remains a gap in dedicated DA techniques designed for the continuous and overlapping loads that dominate industrial energy profiles \cite{angelis2022nilm}.

\rebuttal{Existing industrial NILM datasets present limitations that underscore the need for a more comprehensive resource. For instance, the IMDELD \cite{cg5v-dk02-18} was acquired in a poultry feed factory, covering data over a relatively short duration of 111 days from a single geographic location. Similarly, the HIPE \cite{ehrmann_introducing_2022} was collected in an electronics manufacturing facility over approximately three months with a 5-second sampling interval. Although IMDELD and HIPE reflects certain aspects of manufacturing processes, their short recording period and confinement to a single location prevent it from representing seasonal variations and multi-site operational differences. Moreover, the LILACD \cite{articleDATASET} is derived from a controlled laboratory environment. the dataset does not capture the full range of temporal variability—such as seasonal variations, different load profile from different across multiple industrial facilities and locations.} 

\rebuttal{Motivated by these gaps, we introduce SIDED, a novel open-source industrial dataset generated via Digital Twin simulations, encompassing diverse appliance types across multiple geographical locations and facility profiles. To further enhance data diversity without sacrificing realism, we propose Appliance-Modulated Data Augmentation (\texttt{AMDA}), a computationally efficient DA method that scales each appliance’s power consumption according to its relative load contribution. By preserving signal integrity, \texttt{AMDA} yields more representative training samples that significantly improve NILM performance in the complex industrial environment.}

%%%%%%%%%%%%%%%%%%%%%%%%%%%%%%%%%%%%%%%%%%%%%%%%%%%%%%%%%%%%%%%%%%%%%%%%%%%%%%%%%%%%%%%%%%%%%%%%%%%%%%%%%%%%%%%%%%%%%%%%%%%%%%%%%%%%%%%%%%%%%%%%%%%%%%%%%%%%%%%%%%%%%%%%%%%%%%%%
\rebuttal{\section{Comparative Analysis of Industrial vs. Residential NILM}
\label{sec:comparison}
NILM techniques in residential and industrial settings face very different challenges due to variations in load behavior, signal characteristics, and environmental influences. 
In residential settings, the power consumption is typically generated by a limited number of appliances operating in clearly defined, discrete states. In contrast, industrial settings feature a broad dynamic range with continuously varying loads, overlapping operating periods, and even simultaneous energy production.}

\rebuttal{\textbf{Residential NILM} is generally based on the assumption that the aggregate power consumption, \( x(t) \), can be decomposed into a finite sum of appliance signals that switch between a small number of well-defined states. 
Its mathematical problem formulation is as follows \cite{TANONI2024114703}:
\begin{equation} \label{eq:nilm}
x(t) = \sum_{i=1}^{N} a_i(t) \cdot P_i + \epsilon(t),
\end{equation}
where \(x(t)\) is the aggregate power at time \(t\), \(a_i(t) \in \{0,1\}\) (or drawn from another limited discrete set) represents the state of the \(i^\text{th}\) appliance (e.g., off or on), \(P_i\) is the nominal power consumption of the \(i^\text{th}\) appliance, and \(\epsilon(t)\) models measurement noise.
Residential environments typically involve a few common appliance types that exhibit clear step changes corresponding to switching events. These power profiles are predominantly consumption-only and are characterized by piecewise constant or step-like behavior \cite{10490483}.}

\rebuttal{Instead, \textbf{Industrial NILM} faces a different scenario. 
The different loads operate over a continuous range and often exhibit overlapping and non-discrete patterns due to the complexity of industrial processes. 
Its mathematical model is similar to (\ref{eq:nilm}), but it must accommodate continuous variability and incorporate exogenous factors such as ambient temperature, production schedules, and operational conditions: \cite{TANONI2024114703, LI2023120295}:
\begin{equation}
x(t) = \sum_{i=1}^{N} f_i(\mathbf{u}(t), t) + \eta(t),
\end{equation}
where \(f_i(\mathbf{u}(t), t)\) is a continuous function representing the power profile of the \(i^\text{th}\) load, potentially dependent on external variables \(\mathbf{u}(t)\) (e.g., weather, operational schedules), \(\eta(t)\) captures both measurement noise and unmodeled dynamics.
Industrial systems not only exhibit much higher power levels, with devices such as VFDs and CVAs showing smooth, non-linear power variations, but also include energy production elements like Combined Heat and Power (CHP) units and photovoltaic systems (PV). 
This mixed nature of consumption and production leads to aggregate signals with both positive and negative values. 
Rapid process variations and external influences can cause power fluctuations on the order of 30–50\% over short intervals \cite{YANIV2025115136, 10490483}. 
Moreover, high-quality industrial datasets are often scarce and require advanced preprocessing techniques (e.g., robust normalization, denoising, and augmentation) to deal with the inherent heterogeneity and non-stationarity of the signals.}

\rebuttal{Table~\ref{tab:comparison} summarizes the major differences between residential and industrial NILM. Residential NILM approaches handle discrete, isolated appliance states but fail when applied to industrial settings. In industrial environments, multiple devices and processes are frequently active simultaneously. The resulting overlapping signals create complex aggregate profiles that cannot be neatly decomposed using simple step functions. Industrial power consumption is subject to external factors (e.g., ambient temperature, operational cycles) that have little or no impact in residential settings. Residential models do not account for these covariates. The magnitude and variability of industrial loads, which often include both very high consumption and production components, demand a modeling approach capable of handling a wide range of values and subtle dynamic variations.}

\begin{table}[t]
\centering
\footnotesize
\caption{Key Differences Between Residential and Industrial NILM.}
\label{tab:comparison}
\begin{tabularx}{\columnwidth}{@{}lCC@{}}
\toprule
 & \textbf{Residential NILM} & \textbf{Industrial NILM} \\ \midrule
\textbf{Appliance States}   & Discrete (e.g., off/on)    & Continuous, multi-pattern \\
\textbf{Signal Range}       & Lower power levels         & Higher power levels; includes production \\
\textbf{Data Variability}   & Limited                    & High variability with overlapping signals \\
\textbf{Temporal Dynamics}  & Step-like changes          & Smooth, continuous dynamics \\
\textbf{Noise Characteristics} & Primarily measurement noise & Includes operational variability and external influences \\
\bottomrule
\end{tabularx}
\end{table}

\section{Dataset and Data Augmentation Approach} 
\label{sec:dataset}

\subsection{Synthetic Industrial Dataset for Energy Disaggregation}
% \andrea{We discuss about CVA and VFDs in the intro, then we dont anymore. How does it relate to SIDED?}
The proposed Synthetic Industrial Dataset for Energy Disaggregation (SIDED) is freely available at \url{https://github.com/ChristianInterno/SIDED}, it is compatible with NILMTK \cite{Batra_2014}, a popular open-source toolkit designed to facilitate the comparison and development of NILM algorithms.

\begin{table}[t]
    \centering
        \caption{Parameters characterizing the different facility, which are shared by every location (Germany, US and Japan). 
        % Abbreviations are PV for photovoltaic, CS for cooling system, CHP for combined heat and power, BA for background appliances, and EVSE for electric vehicle supply equipment. 
        % \andrea{ Also add Cooling System size, an info on the EVSE).
        % CHP nominal power cant be MWh... I guess should be something around 180-200kW.}
        }
        % \andrea{check unit measures.} }
    %\small
    \begin{tabularx}{\linewidth}{@{} l *{4}{C} c @{}}
    \toprule
    \textbf{Description} & \textbf{Office} & \textbf{Dealer} & \textbf{Logistic}\\
    \midrule
    Yearly energy demand from the grid ($MWh$) & 1334  & 162 & 1689\\
    % Yearly energy fed into the grid ($MWh$) &  0.182  &  0.375 &  0.650\\
    EVSE yearly energy charging ($MWh$) &  32  & 33 & 32 \\
    PV power ($W$)  per $m^{2}$ & 210 & 150 & 150\\
    PV area ($m^{2}$) & 2000 & 2000  & 10000\\
    CS nominal power ($kW$) & 900 & 250 & 4000\\
    CS Usage Per Year ($h$)& 300 & 300 & 500 \\
    CHP nominal power ($kW$) & 340 & 210 & 900\\
    Yearly energy BA demand ($MWh$) & 2032 & 501 & 4341\\
    % CHP Electrical Efficiency ($\%$) & 35.00 & 35.00 & 35.00 \\
    BA Server power ($kW$) & 69 & 20 & 20\\
    Use server cooling & No & No & Yes\\
    %Storages Capacity ($kWh$) & 0.00 & 2,000.00\% & 0.00\\
    % Battery storage capacity ($kW$) & 0 & 500 & 500\\
%    Battery type & / & Li-Ion & Li-Ion  \\
    Weekly workdays &  5 & 6 & 7\\
    \bottomrule
    \end{tabularx}
\label{tab:Table1}
\end{table}

The dataset contains simulation data for three different realistic industrial facility types: Dealer, Office, and Logistics. 
% \rebuttal{Why realistic? because is a digital twin crafted from an external institution (Ask Sebastian)}
Each facility is defined by its installed appliances and their characteristic electricity consumption profiles. 
In addition to the facility types, each facility can be located in one of three different geographic locations: Offenbach (Germany), Los Angeles (USA), and Tokyo (Japan).
Each location has different environmental conditions defined by the temperature profile and sun radiation, which affect how the photovoltaic, cooling, and heating systems are operated. 
Moreover, working hours and holidays differ between the locations, which manifests itself in different load profiles for each appliance.

\rebuttal{The dataset is generated using a high-fidelity Digital Twin simulator created with a physical simulation library\footnote{[Online] \url{https://www.ea-energie.de/en/projects/green-city-for-simulationx-2/}}. % using Modelica programming language \cite{modelicaspec33}  
The simulation model is based on physical models for all the machinery (CHP, cooling, heating, ventilation...), incorporating the realistic controller modules as in the real devices. 
It models physical effects on the facility buildings, such as irradiation heating, heat diffusion, and energy dissipation in pipes. 
The electricity and heating/cooling demand is taken from real measurement data and adjusted to the facility size.
The weather data are collected through a weather station installed in the three locations for the year 2020.
The simulation was thoroughly calibrated against real world measurement data from an industrial facility for one specific configuration (Office in Offenbach, Germany)~\cite{pub4012,engelMonitorData2025,engelMonitorDataPaper2025}. 
The accumulated discrepancy error over one year between simulation and real-world measurement for this calibrated configuration is below 3\% \cite{pub4012}. 
Due to the fact that the simulator is based on physical models and realistic controller systems, we are guaranteed to produce realistic and physically consistent data also for the other non-calibrated configurations.   
}

%Due to proprietary restrictions associated with the simulation software used to generate the data, we are unable to share the simulator itself, but we are releasing the generated dataset SIDED to support the research community and advance the development of industrial NILM solutions

The parameters characterizing the three simulated facilities simulated are summarized in Table \ref{tab:Table1}. 
The yearly energy demand from the grid for the three building types varies significantly, with the Office requiring 1334 MWh, Dealer requiring 162 MWh, and the Logistic building requiring 1689MWh. 
Each facility in the dataset contains profiles for five different appliances: Electric Vehicle Supply Equipment (EVSE), Cooling Systems (CS), Photovoltaic (PV), Combined Heat and Power (CHP), and Background Appliances (BA), which sums up all smaller devices such as computers, various machinery, servers, or other household appliances.
The EVSE, BA, and SC are consumers of electrical energy, while PV and CHP are producers.
The CHP system uses natural gas to produce electricity and heating, and its efficiency is 35\%.
Important configuration aspects are the size of the IT components (server power), the presence and size of separate server cooling, the presence and size of stationary battery storage, or the working days per week.

\begin{figure*}[t]
\centerline{\includegraphics[width=\linewidth]{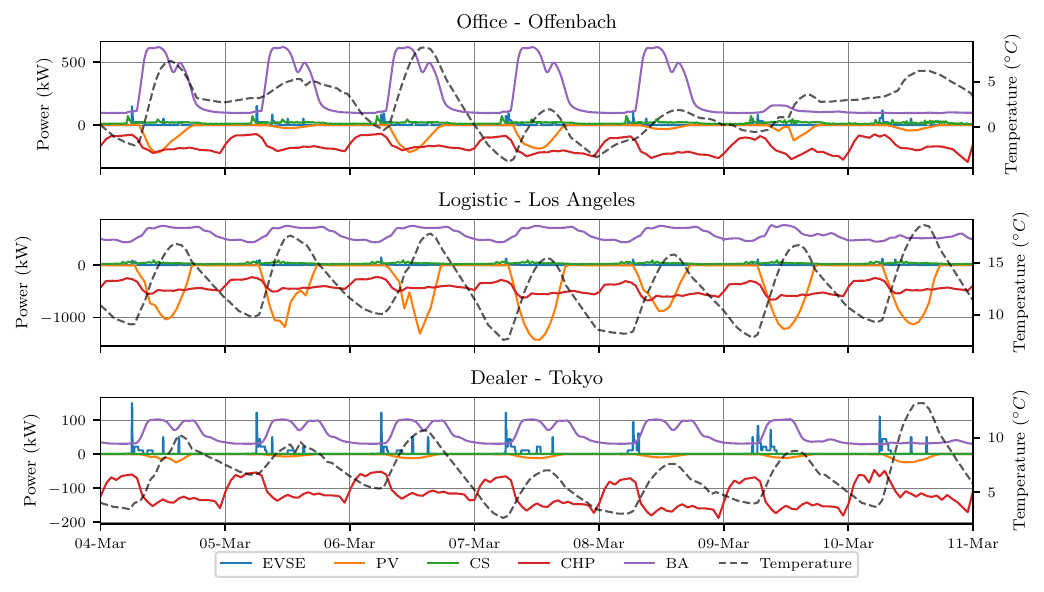}}
\caption{\rebuttal{Example week of SIDED with the different locations and configurations: Office in Offenbach  (top) , Logistic in Los Angeles (center), Dealer in Tokyo (bottom). }
The different appliances are reported: Electric Vehicle Supply Equipment (EVSE), Cooling Systems (CS), Photovoltaic (PV), Combined Heat and Power (CHP), Background Appliances (BA), and ambient temperature. Negative values indicate power generation (PV, CHP).}
\label{fig:fig1}
\end{figure*}

\rebuttal{An example week of SIDED is depicted in Fig.~\ref{fig:fig1}, it shows the three different geolocations (Offenbach, LA, and Tokyo) and the three configurations (Office, Logistic, and Dealer).}
It is possible to observe the different consumption behavior of the different types of appliances and note how the producers (PV and CHP) report a negative power value. 
Additionally, the ambient temperature variation throughout the week is displayed, highlighting its potential correlation with the load profile of certain appliances, e.g., the CHP, CS, and PV.
The aggregate signal $A_t$ is the sum of the other appliances: $A_t = \sum_{i=1}^{n} x_{i,t}$ where $x_{i,t}$ is the signal from the $i^{th}$ appliance at time $t$, and $n$ is the total number of appliances.
This aggregated signal represents what is measured at the main electrical meter i.e., the power drawn from the grid.

\begin{figure}[t]
\centerline{\includegraphics[width=\linewidth]{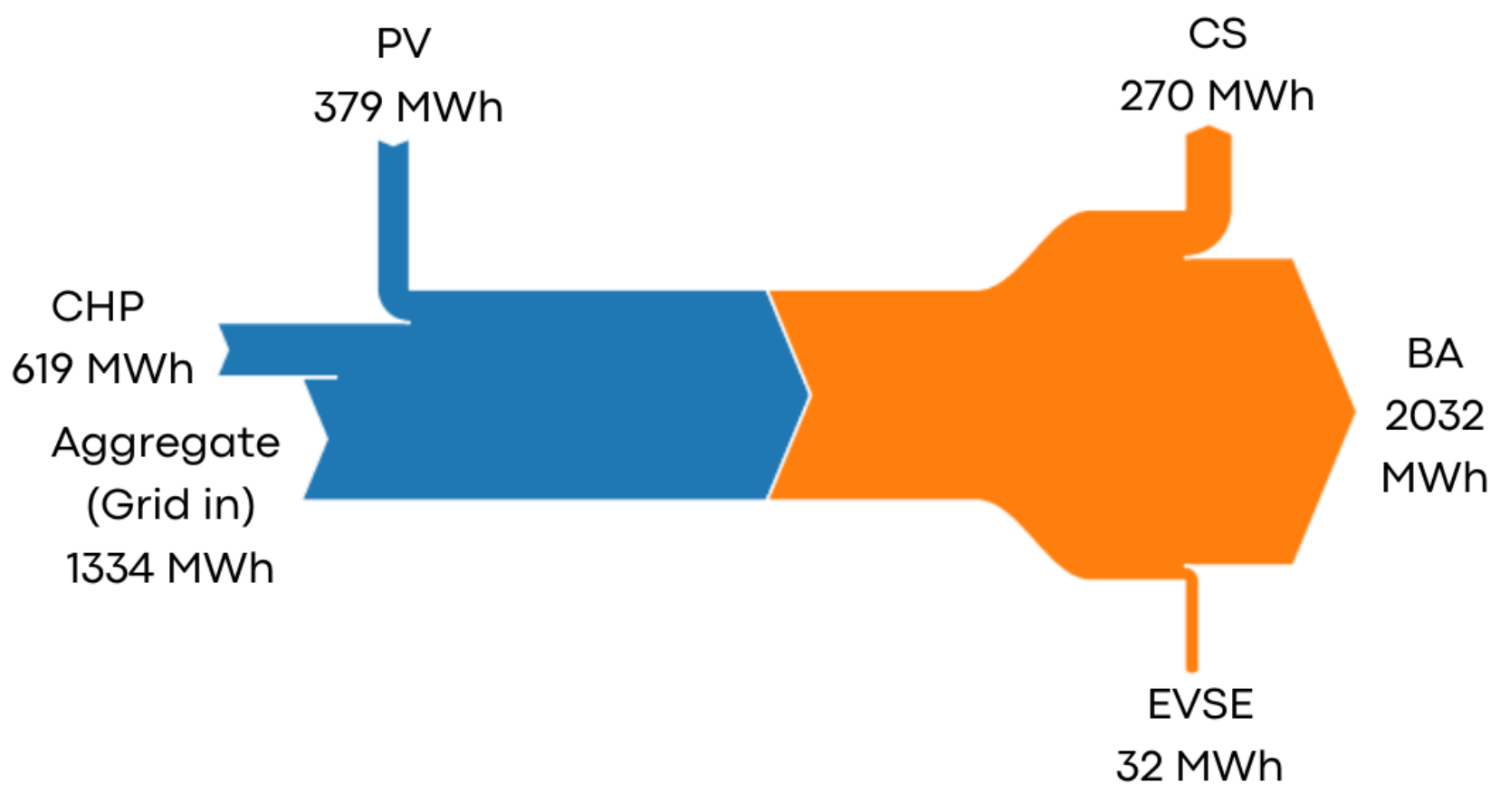}}

\caption{Energy flow accumulated over one year with sources and consumers for the Office configuration in the Offenbach location.
}
\label{fig:fig2}
\end{figure}

Fig.~\ref{fig:fig2} depicts the energy flow diagram for the Office configuration in Offenbach, showing the yearly energy overview. The facility has a total production of 998 MWh (from PV and CHP) and a consumption of 2334 MWh, resulting in 1334 MWh drawn from the distribution grid (referred to as 'Grid in' in the diagram).
% of the energy flow for Office in Offenbach shown in Fig.~\ref{fig:fig2}. 
%It shows how the energy flows from the producers to the consumers' appliances.
It shows the energy yearly overview, with the total production of 998 MWh, and consumption of 2334 MWh, thus 1334 MWh drawn from the distribution grid.

The SIDED dataset consists of the time series of the appliances, and the environment variables for each of the three facilities of each type in each of the three geographic location, in total 9 different configurations. 
The time series of each configuration covers a complete year from January to December, and the measurement readings are taken at a one-minute interval (sampling rate of $\tfrac{1}{60} Hz$). 
As a result, each of the nine simulated facility patterns contains 525,600 individual data points.
For each configuration, the time series included are: ambient temperature ($K$), diffuse sky radiation ($W/m^2$) that the PV system obtains scattered by atmospheric constituents such as clouds and dust, direct radiation ($W/m^2$) from the sun to the PV system, aggregate electric power ($W$), EVSE electric power ($W$), PV production power ($W$), CS electric power  ($W$), CHP electric power ($W$), BA electric power ($W$).
All power values refer to the real power of the appliance. 
The weather and radiation data are not used in this work, but they are included in the SIDED for completeness and future applications.

To justify the choice of configurations and simulated appliances in SIDED, we align with previous classifications of appliances as delineated in~\cite{IQBAL2021106921}. Specifically, SIDED includes both CVA and VFD appliances, which are prevalent in industrial settings and contribute significantly to energy consumption patterns.
% \andrea{Maybe here recall CVA and VFD}
The SIDED dataset incorporates all the following six groups of industrial appliances, each reflecting unique patterns of energy consumption:
\begin{itemize}
\item \textbf{\textit{Constantly-on Variable Appliances}:} Industrial appliances that are never off and always present energy consumption during time, such as the CHP.
\item \textbf{\textit{Periodical}:} Industrial appliances such as EVSE and PV show periodic behavior in energy consumption. 
\item \textbf{\textit{Seasonal}:} Appliances such as PV, CHP, and CS show seasonal behavior in energy consumption, being influenced by the changing weather of the seasons.
\item  \textbf{\textit{Multi pattern}:} Industrial appliances in the multi-pattern category offer multimodal operation with different patterns of energy and power consumption. In SIDED all five appliances can be grouped as multi-pattern.
\item \textbf{\textit{Consumers}:} Industrial appliances that actively consume energy, such as EVSE, CS, and BA.
\item \textbf{\textit{Producers}:} Industrial appliances that actively produce energy, such as PV and CHP.
\end{itemize}

By including these diverse appliance types across various facility configurations and geographic locations, SIDED provides a comprehensive dataset that reflects the complexities of industrial energy consumption. To the best of our knowledge, SIDED is the only industrial dataset that encompasses all these types of appliances, making it a valuable resource for advancing research in industrial NILM.

\subsection{Appliance-Modulated Data Augmentation (\texttt{AMDA})}\label{sec:DA}

Although the SIDED data set contains power signals for nine different facilities, totaling nine years of data, purely data-driven NILM approaches based on DNNs require more variations to generalize well to unseen configurations. Existing DA techniques often fail to capture the unique characteristics of industrial appliances with highly varying power magnitudes. To address this limitation, we introduce \texttt{AMDA}, a computationally efficient DA method designed to enhance the generalization capabilities of NILM models for industrial applications.

\texttt{AMDA} generates signals with greater variance in appliance power levels by scaling each appliance's signal based on its relative contribution to the total power consumption. Instead of applying random scaling factors that may create unrealistic scenarios, we modulate the scaling factor \( S_i \) for each appliance \( i \) using its relative contribution \( p_i \).
The relative contribution \( p_i \) is defined as:
\begin{equation}
p_i = \frac{P_{\text{total},i}}{P_{\text{total}}},
\label{eq:relative_contribution}
\end{equation}
where \( P_{\text{total},i} = \sum_{t=1}^T |x_{i,t}| \) is the sum of the absolute values of the power consumed or produced by appliance \( i \), and \( P_{\text{total}} = \sum_{i} P_{\text{total},i} \) is the total absolute consumption of all appliances.

In order to create additional data samples, \texttt{AMDA} re-scales each signal with an individual scaling factor \( S_i \), 
\begin{equation}
\tilde{x}_{i,t} = S_i \cdot x_{i,t}
\label{eq:rescaled_signal}
\end{equation}
which is  calculated as:
\begin{equation}
S_i = s \cdot (1 - p_i),
\label{eq:scaling_factor}
\end{equation}
where \( s \geq 0 \) is a hyper-parameter controlling the degree of augmentation. 
As a rule of thumb for tuning \( s \), choose values that ensure the scaled signals remain within realistic operational limits for each appliance.
%This can be achieved by defining a range for \( s \) that introduces sufficient variability without distorting the natural consumption profiles of appliances. 
For example, selecting \( s \) values between 1 and 5 allows for meaningful augmentation while respecting appropriate bounds. 
Within this range, \( s \) can be randomly selected to generate diverse augmented samples.
\rebuttal{Importantly, \texttt{AMDA} only modulates the relative magnitudes of individual appliance signals.
It preserves the total aggregate power, ensuring that the overall profiles remain within realistic ranges, a critical property for industrial applications.}

By scaling appliances based on their relative contributions, we amplify the representation of less significant appliances while reducing the impact of those with higher consumption. This ensures that no single appliance dominates the training data, allowing NILM models to learn more balanced patterns and improving their ability to generalize to new configurations.

For example, consider an appliance with a relative contribution \( p_i = 0.25 \) and a chosen scaling parameter \( s = 1.5 \). The scaling factor becomes \( S_i = 1.5 \times (1 - 0.25) = 1.125 \), meaning the appliance's signal is scaled up by 12.5\%.

\begin{figure}[t]
\centering
\includegraphics[width=1\columnwidth]{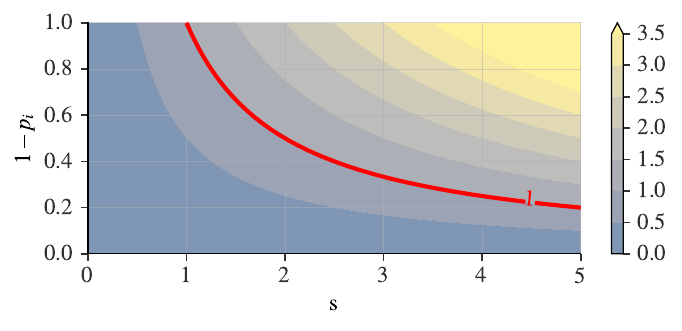}
\caption{Scaling factor \( S_i \) if Eq.~\eqref{eq:rescaled_signal} as a function of \( s \) and \( p_i \). The red line indicates \( S_i = 1 \).}
\label{fig:xy}
\end{figure}

Figure~\ref{fig:xy} illustrates how \( S_i \) varies with different values of \( s \) and \( p_i \). The red line represents \( S_i = 1 \), indicating no scaling. Values below the red line correspond to \( S_i < 1 \), where the appliance's power is reduced.
%The individual scaling factors are applied to rescale each appliance's signal:
%\begin{equation}
%\tilde{x}_{i,t} = S_i \cdot x_{i,t}.
%\label{eq:rescaled_signal}
%\end{equation}
To generate one set of re-scaled data samples for augmentation, the procedure is applied to the series in the dataset with a fixed value of $s$. Algorithm~\ref{alg:amda} summarizes the \texttt{AMDA} method.

\begin{algorithm}[t]
\small
\caption{Appliance-Modulated Data Augmentation (\texttt{AMDA})}
\begin{algorithmic}[1]
\Procedure{\texttt{AMDA}}{Dataset, $s$}
    \For{each appliance \( i \) in Dataset}
        \State Compute \( p_{i} \leftarrow \dfrac{P_{\text{total},i}}{P_{\text{total}}} \) \Comment{Relative contribution}
        \State Compute \( S_i \leftarrow s \cdot (1 - p_{i}) \) \Comment{Scaling factor}
        \State Scale signal \( \tilde{x}_{i,t} \leftarrow S_i \cdot x_{i,t} \)
    \EndFor
    \State Merge all \( \tilde{x}_{i,t} \) to form augmented aggregate signal
    \State Combine augmented signals with the original dataset 
\EndProcedure
\end{algorithmic}
\label{alg:amda}
\end{algorithm}

Table~\ref{tab:relative_contributions} presents the relative contributions \( p_i \) and total consumption \( P_{\text{total},i} \) for each appliance in different facility types located in Offenbach. Appliances like BA (Building Automation) have high \( p_i \), resulting in lower scaling factors and reduced contributions in the augmented data. Conversely, appliances with lower \( p_i \), such as PV and EVSE, receive higher scaling factors, enhancing their representation.

\begin{table}[t]
\centering
\caption{Relative contributions \( p_{i} \) and total consumption \( P_{\text{total},i} \) (in kW) for each appliance in different facility types in Offenbach.}
\begin{tabularx}{\columnwidth}{@{} l *{6}{C} @{}}
\toprule                 
& \multicolumn{2}{c}{\textbf{Office}} & 
\multicolumn{2}{c}{\textbf{Dealer}} & 
\multicolumn{2}{c}{\textbf{Logistics}}  \\
\cmidrule(r){2-3} \cmidrule(r){4-5} \cmidrule{6-7}
& \( p_i \) & \( P_{\text{total},i} \) & \( p_i \) & \( P_{\text{total},i} \) & \( p_i \) & \( P_{\text{total},i} \) \\
\midrule     
EVSE & 0.009 & 1,962 & 0.035 & 1,986 & 0.004 & 1,967 \\
PV & 0.114 & 22,800 & 0.041 & 2,303 & 0.181 & 92,626 \\
CS & 0.080 & 16,200 & 0.018 & 1,020 & 0.088 & 45,000 \\
CHP & 0.186 & 37,185 & 0.373 & 21,009 & 0.221 & 113,446 \\
BA & 0.601 & 121,921 & 0.533 & 30,076 & 0.507 & 260,497 \\
\bottomrule
\end{tabularx}
\label{tab:relative_contributions}
\end{table}

By generating new proportions between the contributions of each appliance, \texttt{AMDA} efficiently creates a training set with expanded and realistic variance, without relying on random modifications. 
This method maintains the temporal coherence of the signals and avoids introducing artifacts that could mislead the learning algorithms.
\rebuttal{It should be noted that the computational demand of the \texttt{AMDA} method itself scales linearly with the amount of additional data generated. The operations to generate an additional data series are very efficiently implementable as they only consist of copy and rescale operations.}

%In summary, \texttt{AMDA} offers a simple and effective approach to augmenting industrial NILM datasets. By adjusting appliance contributions in a controlled manner, it enhances the NILM model's ability to generalize to unseen configurations with varying appliance behaviors.

\section{Experimental Setup}\label{sec:exp}
\rebuttal{The code and data used in this work are publicly available at \url{https://github.com/ChristianInterno/SIDED}.}

\subsection{Deep Learning Models}

We conduct experiments with the deep neural network (DNN) architectures Long Short-Term Memory (LSTM)from \cite{10044076}, Temporal Convolutional Neural Network (TCN), and Attention-TCN (ATCN) from \cite{9713379}. 
We employ the single-appliance NILM setting where the input is the aggregated power signal and the model extracts the power of a single appliance e.g., the CHP.
In all of our experiments, we use the sequence-to-point network \cite{10.5555/3504035.3504353} structure: given a time series of aggregate power readings $X = \{x_1, x_2, ..., x_T\}$, we use a sliding window of size $w$ to create input sequences $X_t = \{x_{t-w/2}, ..., x_t, ..., x_{t+w/2}\}$ for each time point $t$ with $t > w/2$ and $t \leq T - w/2$. 
The target output is $y_t$, the power reading of the appliance at time $t$. The DNN learns a function $F$ mapping $X_t$ to $y_t$:
$F(X_t; \theta) = y_t$,
where $\theta$ represents the DNN parameters.  

We use an LSTM network with three hidden layers, each containing 128 units. LSTMs are well-suited for capturing temporal dependencies in sequential data, making them effective for NILM tasks.
The TCN model comprises eight layers that combine temporal convolutional and pooling operations. It has a receptive field covering the entire input sequence, allowing it to model long-range temporal patterns. The network processes single-channel input, produces 128 output channels, and incorporates batch normalization, ReLU activation, and a dropout rate of 0.33 to prevent overfitting.
The ATCN~\cite{9713379} extends the TCN by introducing attention mechanisms. It generates multiple temporal residual blocks and uses attention to highlight the importance of each block for the target appliance, effectively capturing both long-term and short-term dependencies.
All models are trained using the Adam optimizer with a learning rate of 0.001. We set the batch size to 64, employing early stopping based on validation loss to prevent overfitting.

% In our experiments, we used an LSTM with three hidden layers with 128 hidden units each. 
% The TCN includes eight layers combining temporal convolutional and pooling layers, with a receptive field that covers entirely the input sequence. 
% It takes a single-channel input, produces 128 output channels, and applies batch-normalization, ReLU activation, and a 0.33 dropout rate to prevent overfitting. 
% The proposed ATCN from \cite{9713379} adapts the TCN structure by generating different temporal residual blocks and incorporating an attention mechanism to highlight the importance of these blocks for each appliance, thereby incorporating long-term and short-term dependencies from various appliances.

\subsection{Data Preprocessing and Evaluation Metrics \label{sec:secpre}}
For the current application, the one-minute resolution of the original data in the SIDED dataset is not necessary and we re-sample all time series to a sampling period of $T_s = 5$ min in order to reduce the computation effort.
Before feeding the data to any model, we normalize it for each building configuration with a RobustScaler \cite{scikit-learn}, which normalizes data by subtracting the median and then scaling it according to the interquartile range. This ensures scaling is specific to each configuration, capturing the unique power distribution of appliances and avoiding inconsistencies from using a universal scaler across varying configurations.

% Mathematically, this can be expressed as:
% \(
% x_{\text{scaled}} = \frac{x - \text{median}(x)}{\text{IQR}(x)}
% \)
% where \( x \) represents an original data point, \(\text{median}(x)\) is the median of all data points in the feature, and \(\text{IQR}(x) = \text{Q3}(x) - \text{Q1}(x)\), with \(\text{Q1}\) and \(\text{Q3}\) being the first and third quartiles, respectively. 

Although our data is simulated, the signals in the SIDED dataset have varying power levels and can exhibit extreme values due to the dynamics of industrial appliances. We found that the RobustScaler effectively handles these variations and ensures that the model is not heavily influenced by extreme values.
In order to be able to compare the results, all model outputs are de-normalized again in order to get time series signals on the original scales, and all performance metrics described below are calculated with de-normalized data.

We use sliding windows with a size of one day (288 samples at 5-minute intervals) and a stride of 25 minutes (5 samples). The window size is a hyper-parameter chosen to naturally reflect the daily characteristic behavior of the appliances, allowing the models to learn specific behaviors in the data.

As an example, Fig.~\ref{fig:fig1} shows the usage cycles over a week of all signals for one Office Offenbach facility configuration. It highlights the periodic nature of device usage and the significant impact of working days on the BA signal. Additionally, the temperature varies throughout the week. This fluctuation may have a short-term correlation with the power consumption of specific appliances such as CHP, EVSE, and PV, and a long-term correlation with CS.

In the context of NILM, various metrics exist for evaluating results~\cite{IQBAL2021106921}. Since we are dealing with regression tasks and industrial appliances that are continuously on, rather than household appliances with on-off patterns, we employ regression-based evaluation metrics. Specifically, we use Mean Absolute Error (MAE), Mean Squared Error (MSE), the coefficient of determination ($R^2$), and the Normalized Disaggregation Error (NDE) \cite{9087706}. The NDE metric provides a dimensionless and normalized measure of disaggregation quality, facilitating fair comparisons across different appliances.
\rebuttal{These four metrics collectively offer a well-rounded evaluation of the complex and dynamic nature of industrial signals.}

\section{Experiments and Results}
\label{sec:experiment}
We demonstrate the benefits of the proposed \texttt{AMDA} method in enhancing model generalization capabilities in realistic industrial settings with SIDED. We use quantitative metrics, distance measures to provide a comprehensive evaluation on:

A. \textbf{Appliance Variation Scenario:} \newrebuttal{As a "proof of concept" to show the effectiveness of \texttt{AMDA} we simulate a scenario where the power level of an appliance in the facility changes after the model has been trained. 
This experiment proximate situations where new appliances are introduced, or existing ones are removed or reconfigured.}

B. \textbf{Facility Variation Scenario:} We examine the performance of the NILM system when applied to a facility configuration different from the one used during training. 
This includes changes in geographic location, load profiles, and power levels for each appliance. We also compare training times and the computational effort required for \texttt{AMDA}.

\newrebuttal{C. \textbf{Ablation Study on Hyperparameter \textit{s}:} An ablation study on the hyperparameter $s$ to analyze its impact on performance. This provides empirical guidance for tuning the \texttt{AMDA} method and demonstrates its robustness across a range of values.}

E. \textbf{Data Distribution Alignment Analysis}: Finally, we quantitatively evaluate the information gained via \texttt{AMDA}, via KL and JS divergence to quantify the alignment between the distributions of the training and testing data. We use dimensionality reduction techniques like UMAP~\cite{umap2018} to visualize the data distributions. This approach allows us to demonstrate how \texttt{AMDA} brings the training data distribution closer to that of the testing data, improving model performance.

In all experiments, we focus on single-appliance NILM, specifically targeting the disaggregation of the CHP appliance from the aggregate signal. We chose the CHP appliance due to its unique characteristics, including seasonality, susceptibility to weather variables, large fluctuations in machinery power, and its multi-pattern behavior.
These factors make the CHP a challenging target for load disaggregation \cite{9179030, DBLP:journals/corr/abs-2105-00349}.
\rebuttal{The CHP system output not only vary in amplitude but also in their operating modes (e.g., duration, shape of transitions, and temporal dynamics) across different facilities due to their dependence on facility background load, demand and external temperatures. 
Due to the inclusion of different facility types and locations in our experimental setup (described below in Section \ref{sec:FVS}) we can test the proposed augmentation algorithm in out-of-distribution scenarios, since the augmentation cannot natively adjust for different operating modes. 
For instance, by using an entirely unseen facility configuration (Logistics) during testing. The results indicate that, even under these conditions, models trained with \texttt{AMDA} outperform those trained with conventional augmentation methods.}

To ensure a statistically robust evaluation of the models' performance, we repeat each experiment 20 times with different random seeds and model weight initializations. To assess the statistical significance of our results, we use the Critical Difference plot \cite{JMLR:v7:demsar06a} to visualize and compare the performance of the models.

\subsection{Appliance Variation Scenario}

\newrebuttal{This experiment evaluates how our \texttt{AMDA} method improves a model's adaptability and generalization when the power consumption of a dominant appliance changes. We partitioned the Office Offenbach dataset into training (72.25\%), validation (12.75\%), and test (15\%) sets. The BA appliance, which accounts for over 60\% of the total energy consumption, was specifically modified in our test sets to simulate realistic changes, such as replacing or removing a major appliance.}

\newrebuttal{To achieve this, we scaled the BA appliance's power signal in the test set by a factor $s$, ranging from 0 (appliance removed) to 2 (power doubled), while all other appliances remained unchanged. The baseline scenario, identical to the training data, is represented by $s=1$. We then compared LSTM, TCN, and ATCN models trained with and without \texttt{AMDA}. The augmented training set for \texttt{AMDA} was generated using scaling hyperparameters $s_1=1.5$ and $s_2=4$. As designed, \texttt{AMDA}'s scaling factors are anti-correlated with an appliance's relative contribution; this reduces the influence of dominant appliances (like BA) and increases the influence of minor appliances, creating a more diverse training set.}

\begin{figure}[t]
    \centerline{\includegraphics[width=\linewidth]{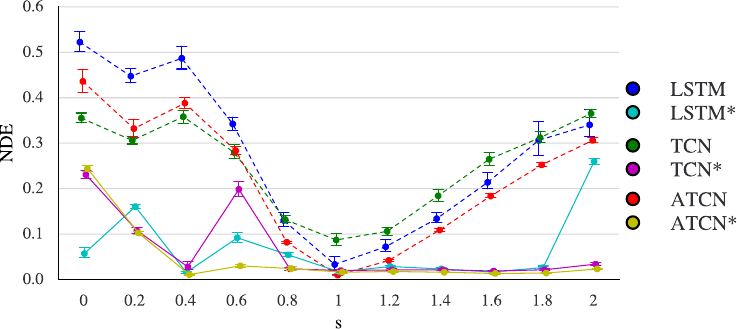}}
    \caption{NILM performance measured by NDE for DNN models trained on original and \texttt{AMDA}-augmented datasets (indicated by *) and tested on artificially created test sets with varying scaling factors $s$ applied to the BA appliance for the appliance variance scenario (see main text).}
    \label{fig:fig4}
\end{figure}

\newrebuttal{The results clearly demonstrate \texttt{AMDA}'s effectiveness. As shown in Table~\ref{tab:tab3}, models augmented with \texttt{AMDA} significantly outperformed their non-augmented counterparts, achieving a consistently lower average rank across all metrics. The ATCN model with \texttt{AMDA} was the top performer, improving its MAE by $\sim$60\%, MSE by $\sim$83\%, $R^2$ by $\sim$54\%, and NDE by $\sim$81\%. Figure~\ref{fig:fig4} further shows that \texttt{AMDA}-enhanced models maintain superior performance across all scaling factors ($s$). We did, however, observe a performance decrease for all models when the BA signal was significantly reduced ($s < 1$). This is expected, as diminishing the main contributor to the aggregate signal reduces the signal's overall variance, making the disaggregation task inherently more difficult.}

\begin{table}[t]
    \centering
    \caption{Average error metrics for experiment 'Appliance Variation Scenario'. Asterisks (*) indicate models trained with \texttt{AMDA}. The average rank is calculated by ranking each model based on its performance across all metrics.}
    \label{tab:tab3}
    \scriptsize
    \begin{tabularx}{\columnwidth}{@{}l *{6}{C}@{}}
        \toprule
        & \textbf{LSTM} & \textbf{TCN} & \textbf{ATCN} & \textbf{LSTM*} & \textbf{TCN*} & \textbf{ATCN*} \\
        \midrule
        MAE [MW] & $0.033$ & $0.033$ & $0.029$ & $0.015$ & $0.013$ & $\bm{0.012}$ \\
        & $\pm0.012$ & $\pm0.008$ & $\pm0.012$ & $\pm0.008$ & $\pm0.009$ & $\pm0.008$ \\
        MSE [$(\text{MW})^2$] & $2.801$ & $2.564$ & $2.247$ & $0.696$ & $0.672$ & $\bm{0.476}$ \\
        & $\pm1.728$ & $\pm1.052$ & $\pm1.427$ & $\pm0.761$ & $\pm0.776$ & $\pm0.689$ \\
        $R^2$ & $0.490$ & $0.540$ & $0.596$ & $0.879$ & $0.880$ & $\bm{0.911}$ \\
        & $\pm0.318$ & $\pm0.183$ & $\pm0.254$ & $\pm0.128$ & $\pm0.128$ & $\pm0.111$ \\
        NDE & $0.271$ & $0.254$ & $0.223$ & $0.066$ & $0.067$ & $\bm{0.040}$ \\
        & $\pm0.169$ & $\pm0.103$ & $\pm0.140$ & $\pm0.074$ & $\pm0.076$ & $\pm0.067$ \\
        \addlinespace
        Avg. rank & 5.363 & 5.272 & 4.090 & 2.545 & 2.363 & $\bm{1.363}$\\
        \bottomrule
    \end{tabularx}
\end{table}

\subsection{Facility Variation Scenario}
\label{sec:FVS}
\newrebuttal{We evaluate model generalization in a challenging out-of-distribution scenario by training on data from the Office facility in Offenbach (80\% training, 20\% validation) and testing on the entirely unseen data from the Logistics facility in Los Angeles. This setup simulates the real-world deployment of a model in a facility with different load profiles, appliance parameters, and operating conditions. We report results using the best-performing ATCN model from previous experiment.}

\newrebuttal{To assess \texttt{AMDA}'s effectiveness, we compare it against two non-augmented baselines and six SoTA augmentation techniques. The non-augmented baselines are \textbf{Base}, model trained only on the Office Offenbach dataset, and \textbf{Enlarged}, the model trained on a dataset combining the Office and Dealer configurations from Offenbach.}

\newrebuttal{The SoTA augmentation methods for comparison are Random Scaling (RDM) \cite{Delfosse2020DeepLA}, Jitter, Shift, FreqDropout \cite{Wan2024Nonintrusive}, Denton-Cholette Interpolation \cite{LeQuy2022Data}, OFFSETAUG \cite{Francou2023Expanding}, and InterUser \cite{Yang2025Intra}. We apply those SoTA methods to augment the Office Offenbach (Base) dataset. Each method is set to double the size of Base.
For RDM, we generate 14 scaled versions for each appliance's signal, ranging from an 80\% reduction to a 1000\% increase in appliance power. This results in a total of 12,623,040 data points.}

\newrebuttal{We apply our \texttt{AMDA} method (Algorithm~\ref{alg:amda}) to create two augmented datasets: \textbf{Base} (augmenting the Base dataset) and \textbf{Enlarged} (augmenting the Enlarged dataset). The scaling factor hyperparameter s for \texttt{AMDA} was set $s=2.5$, based on the findings of our ablation study, detailed in following Section \ref{sec:ablation}.}

\begin{figure}[t]
    \centerline{\includegraphics[width=\linewidth]{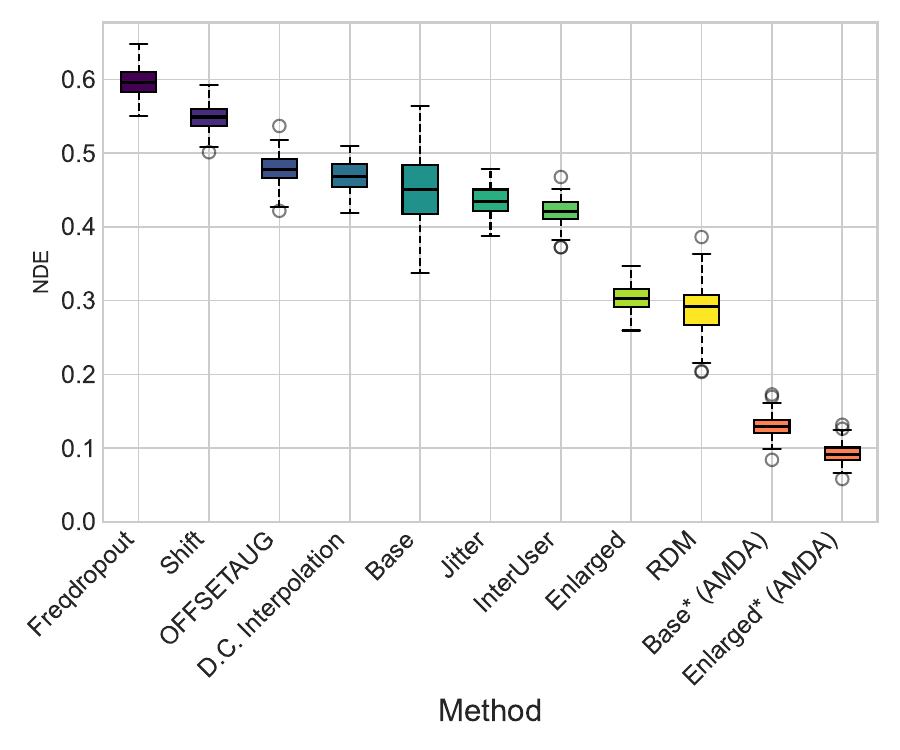}}
    \caption{NILM performance \rebuttal{on the test dataset of a Logistics facility in Los Angeles} with NDE metric in the Facility Variation Scenario for several DA methods and \texttt{AMDA}.}
    \label{fig:fig5}
\end{figure}

\newrebuttal{The results of this comparison are presented in Table~\ref{tab:tab4_adjusted} and the NDE metric is graphically shown in Fig.~\ref{fig:fig5}. The data clearly demonstrates the better performance of the \texttt{AMDA}. The Enlarged*\texttt{AMDA} configuration achieves the best performance across all metrics, with an NDE of 0.167 with the next best method (RDM) achieving an NDE of 0.290.
Notably, \texttt{AMDA} achieves best results with the same data volume and computational cost as the other augmentation techniques, highlighting its augmentation quality. The performance of Enlarged*\texttt{AMDA} also closely approaches the theoretical Maximum-achievable-Performance (MP) benchmark (NDE of 0.009), which was obtained by training the model directly on the test data.}

\newrebuttal{The FreqDropout and Shift data augmentation methods did not improve performance over the Base. SoTA methods showed only marginal improvements, suggesting that DA techniques designed for domestic NILM may not be effective in industrial environments.}

\begin{figure*}[t]
    \centerline{\includegraphics[width=0.8\linewidth]{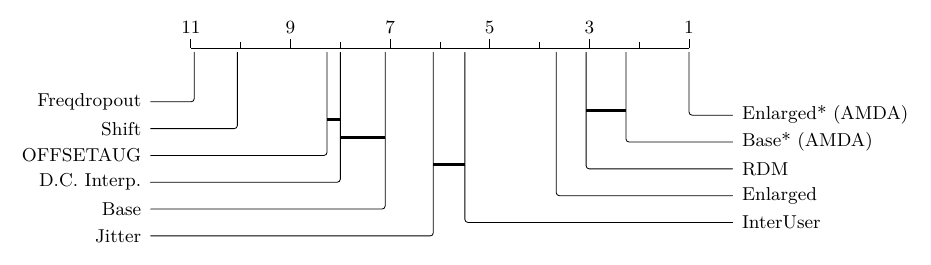}}
    \caption{Critical Difference diagram summarizing Experiment 2. Asterisks (*) indicate datasets augmented with \texttt{AMDA}.}
    \label{fig:fig6}
\end{figure*}

 \newrebuttal{The statistical analysis shown in the Critical Difference plot (Fig.~\ref{fig:fig6}) confirms these findings. The plot, which visualizes the results of a Friedman test followed by a Nemenyi post-hoc test, shows that the \texttt{AMDA}-based methods (\textbf{Base} and \textbf{Enlarged}) are ranked significantly higher than all other SoTA baselines.}

\begin{table*}[t]
    \centering
    \caption{Average results of the 'Facility Variation Scenario' comparing AMDA-augmented models with non-augmented (Base, Enlarged) and SoTA data augmentation methods.  Dataset size increment relatively to Base.}
    \label{tab:tab4_adjusted}
    \scriptsize
    \begin{tabularx}{\textwidth}{@{}l*{11}{C}@{}}
        \toprule
        \textbf{Metric} & \textbf{Base} & \textbf{Enlarged} & \textbf{Freqdropout} & \textbf{Shift} & \textbf{Jitter} & \textbf{InterUser} & \textbf{D.C. Interp.} & \textbf{OFFSETAUG} & \textbf{RDM} & \textbf{Base* (AMDA)} & \textbf{Enlarged* (AMDA)} \\
        \midrule
        \textbf{Dataset size} & 631,152 & 
        100\%&+100\% & +100\% & +100\% & +100\% & +100\%  & +100\% & +1500\% & +100\%& +300\%\\
        \midrule
        \textbf{MAE [MW]} & 0.155 & 0.131 & 0.141 & 0.139 & 0.137 & 0.133 & 0.132 & 0.129 & 0.127 & 0.080 & \textbf{0.069} \\
        & $\pm$.007 & $\pm$\textbf{.011} & $\pm$.011 & $\pm$.010 & $\pm$.009 & $\pm$.008 & $\pm$.005 & $\pm$.013 & $\pm$.012 & $\pm$.006 & $\pm$.006 \\
        \textbf{MSE [MW$^2$]} & 0.048 & 0.032 & 0.042 & 0.040 & 0.038 & 0.034 & 0.033 & 0.031 & 0.030 & 0.013 & 0.010 \\
        & $\pm$.005 & $\pm$.002 & $\pm$.004 & $\pm$.005 & $\pm$.004 & $\pm$.003 & $\pm$.004 & $\pm$.004 & $\pm$.004 & $\pm$.002 & $\pm$.001 \\
        \textbf{R$^2$} & 0.216 & 0.435 & 0.224 & 0.280 & 0.312 & 0.341 & 0.410 & 0.451 & 0.482 & 0.778 & \textbf{0.950} \\
        & $\pm$.092 & $\pm$.051 & $\pm$ .034 & $\pm$.061 & $\pm$.058 & $\pm$.055 & $\pm$.049 & $\pm$.052 & $\pm$.058 & $\pm$.024 & $\pm$.026 \\
        \textbf{NDE} & 0.451 & 0.305 & 0.596 & 0.548 & 0.436 & 0.425 & 0.470 & 0.475  & 0.290 & 0.271 & \textbf{0.167} \\
        & $\pm$.051 & $\pm$.017 & $\pm$.041 & $\pm$.039 & $\pm$.038 & $\pm$.031 & $\pm$.033 & $\pm$.035 & $\pm$.035 & $\pm$.016 & $\pm$.013 \\        
        \bottomrule
    \end{tabularx}
\end{table*}

\subsection{Ablation Study on Hyperparameter \textit{s}}
\label{sec:ablation}
\newrebuttal{To provide empirical guidance on tuning our proposed \texttt{AMDA} method and to evaluate its robustness, we perform an ablation study on its key hyperparameter, \textit{s}. We analyzes how different values of \textit{s} impact model performance in the challenging out-of-distribution "Facility Variation Scenario."}

\newrebuttal{The experimental setup was identical to that in Section VI.B: the ATCN model was trained on the \textbf{Base} dataset (Office, Offenbach) augmented by \texttt{AMDA} using different values of \textit{s}, and then tested on the unseen Logistics facility in Los Angeles. We varied \textit{s} across a range from 1.5 to 5 with a step of 0.25 to observe its effect on disaggregation performance.}

\begin{figure}[t]
\centering
\newrebuttal{
\includegraphics[width=\columnwidth]{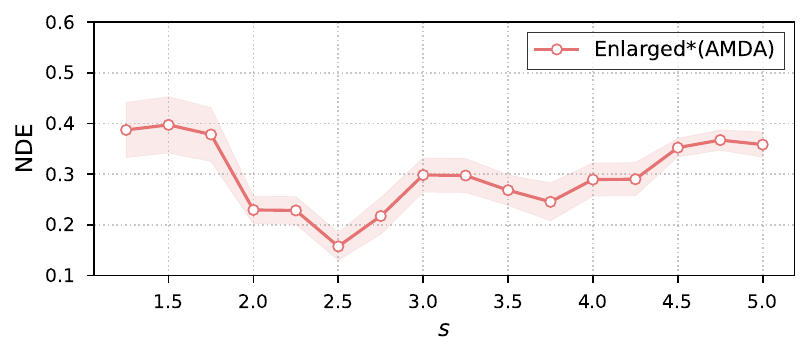}
\caption{NILM performance (NDE) as a function of the \texttt{AMDA} hyperparameter \textit{s}. The results show a clear optimal range for \textit{s} between 1.0 and 5.0, demonstrating the method's robustness.}
\label{fig:ablation_s}
}
\end{figure}

\newrebuttal{The results indicate that \texttt{AMDA} is robust to the choice of \textit{s} within a wide range. As shown in Figure~\ref{fig:ablation_s}, performance improves significantly as \textit{s} increases from 1.5, with the best results achieved for \textit{s=2.5}. This analysis confirms that choosing \textit{s} within the range of [1, 5], as recommended in Section~\ref{sec:DA}, is a practical and effective rule of thumb. It provides substantial performance gains while remaining safely within the zone of optimal, robust performance.}

\subsection{Data Distribution Alignment Analysis}

\begin{figure*}[t]
    \centerline{\includegraphics[width=\linewidth]{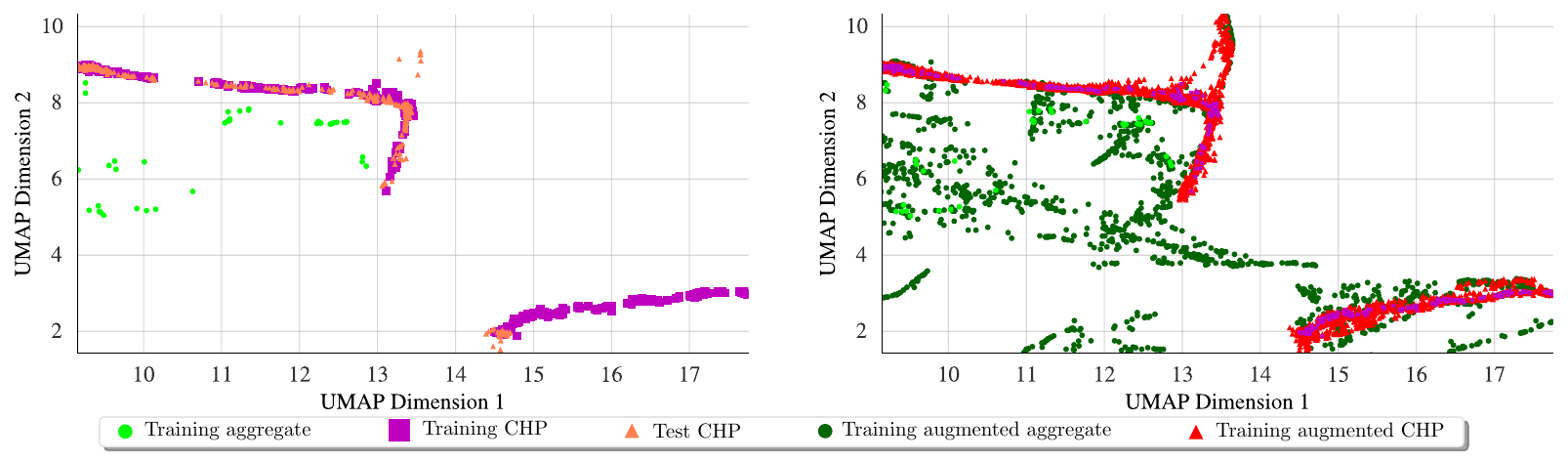}}
    \caption{Detail of two-dimensional UMAP projections of energy signals for the CHP appliance. Left: \textbf{Base} training data and target CHP data. Right: \texttt{AMDA}-augmented \textbf{Base*} data and target CHP data.}
    \label{fig:fig7}
\end{figure*}

We project the time-series training signals of the \textbf{Base} dataset and its \texttt{AMDA}-augmented version Base* onto a two-dimensional space using UMAP, as shown in Figure~\ref{fig:fig7}. Focusing on the CHP target appliance, we provide a zoomed section to emphasize the CHP-related data.

The left panel depicts the Base training data alongside the test CHP data from the Logistics Los Angeles building configuration.
While there is some overlap between the training and test CHP data, a significant portion of the test data lies in regions where the training data is sparse, particularly in the lower right part of the plot. This indicates that models trained on the Base dataset may need to extrapolate in these regions, potentially reducing their accuracy.

In contrast, the right panel shows the \texttt{AMDA}-augmented \textbf{Base*} data. The augmented data covers a larger region of the feature space and extends towards the complete range of the test CHP data. This suggests that \texttt{AMDA} effectively enriches the training data, preserving the structure and neighborhood relations of the data samples through relative scaling. As a result, models trained with \texttt{AMDA}-augmented data are better equipped to generalize to unseen data.

To quantitatively analyze the alignment between the distributions of the CHP data in different training datasets, we compute the KL divergence and JS divergence between each training dataset and the test set, calculated from the original data distributions (not from the UMAP projections), as reported in Table~\ref{tab:divergences}. 
We observe that the \texttt{AMDA}-augmented datasets (Base* and Enlarged*) have significantly lower KL and JS divergence values compared to the other methods. Specifically, the KL divergence for Enlarged* is 0.314, a reduction of approximately 51\% compared to the Base dataset (0.645). The JS divergence for Enlarged* is 0.180, representing a reduction of about 49\% compared to the Base dataset (0.350). This improved alignment is crucial for the model's ability to generalize to unseen data, leading to better performance in energy disaggregation tasks.

\begin{table}[t]
    \centering
    \caption{KL and JS divergence values between each training and the test dataset. Lower values indicate better alignment with the test distribution. Percentages show a reduction compared to the Base dataset.}
    \label{tab:divergences}
    \begin{tabularx}{\columnwidth}{@{} l *{2}{C} | *{2}{C} @{}}
    \toprule
    \textbf{Dataset} & \textbf{KL Div.} & \textbf{\% Red.} & \textbf{JS Div.} & \textbf{\% Red.} \\
    \midrule
    Base & 0.645 & - & 0.350 & - \\
    Enlarged & 0.553 & 14\% & 0.326 & 7\%\\
    RDM & 0.518 & 20\% & 0.297 & 15\% \\
    Base* (\texttt{AMDA}) & 0.335 & 48\% & 0.224 & 36\% \\
    Enlarged* (\texttt{AMDA}) & \textbf{0.314} & \textbf{51\%} & \textbf{0.180} & \textbf{49\%} \\
    \bottomrule
    \end{tabularx}
\end{table}

\section{Conclusions and Future Work}
\label{sec:conclusion}

We addressed the central challenge of data scarcity in purely data-driven industrial Non-Intrusive Load Monitoring (NILM) approaches, by introducing the Synthetic Industrial Dataset for Energy Disaggregation (SIDED). 
SIDED is a high-quality dataset of electricity consumption and production profiles for industrial facilities, generated using a Digital Twin. It encompasses three different types of facilities across three geographic locations, providing diverse scenarios for NILM research.
The configuration possibilities in industrial facilities, due to the variety in types and sizes of appliances, are much greater than in typical household settings. 
Consequently, generating datasets that cover a wide range of possible configurations is often not feasible, which hinders the generalization capabilities of machine learning-based NILM models. To address this, we also introduced the Appliance-Modulated Data Augmentation (\texttt{AMDA}) method, which generates a large amount of training data with increased variability by scaling appliance contributions based on their relative impact in a simple and computationally efficient manner.
Our experiments demonstrated that \texttt{AMDA} enhances the \rebuttal{generalization}of NILM models in industrial settings using the SIDED dataset. Models trained with \texttt{AMDA} effectively disaggregated signals even when primary appliances were scaled to out-of-distribution sizes in the test data, simulating changes in appliance behavior. Additionally, we assessed the generalization capabilities of NILM models trained with \texttt{AMDA} on unseen building configurations. The results showed that augmenting the training data with \texttt{AMDA} led to significant performance improvements, surpassing those achieved with conventional data augmentation methods.

While SIDED provides a valuable resource for industrial NILM research, it has limitations.  Factors such as unlabeled appliance data and measurement noise present in actual facilities might not be fully represented. 
Regarding the \texttt{AMDA} method, its effectiveness relies on the assumption that scaling appliance contributions based on their relative impact can capture the variability seen in different industrial scenarios. However, \texttt{AMDA} assumes that the scaling of appliances does not lead to unrealistic operational states given specific domain knowledge. Therefore, care must be taken to ensure that the augmented data remains within realistic bounds.

\rebuttal{Expanding SIDED with additional configurations and appliance types can further enhance its utility.} \newrebuttal{A promising area for future work is to explore variations of AMDA that use a sliding window technique to account for temporal information, allowing the scaling factor to be tuned per window.}

Integrating the SIDED dataset and \texttt{AMDA} into online learning scenarios or federated learning settings with non-IID data \cite{10651455} and constrained resources environments \cite{interno2024adaptive} can be explored.

\section*{Acknowledgment}
Christian Internò acknowledges funding from the Honda
Research Institute Europe.

\bibliographystyle{IEEEtran}
\bibliography{bibliography}

\end{document}